\documentclass{article}

\PassOptionsToPackage{numbers, compress}{natbib}

  \usepackage[preprint]{neurips_2020}

\usepackage[utf8]{inputenc} 
\usepackage[T1]{fontenc}    
\usepackage{hyperref}       
\usepackage{url}            
\usepackage{booktabs}       
\usepackage{amsfonts}       
\usepackage{nicefrac}       
\usepackage{microtype}      

\usepackage{amsmath}
\usepackage{graphicx}
\usepackage{wrapfig}
\usepackage{caption,subcaption}
\usepackage{booktabs}
\usepackage{multirow}
\usepackage{tikz}
\usepackage{enumitem}

\title{Neuralizing Efficient Higher-order Belief Propagation}

\author{%
	Mohammed Haroon Dupty\thanks{corresponding author} \\
	School of Computing\\
	National University of Singapore\\
	\texttt{dmharoon@comp.nus.edu.sg} \\
	\And
	Wee Sun Lee \\
	School of Computing\\
	National University of Singapore\\
	\texttt{leews@comp.nus.edu.sg} \\
}

\begin{document}
	
	\maketitle
	
	\begin{abstract}
		Graph neural network models have been extensively used to learn node representations for graph structured data in an end-to-end setting. These models often rely on localized first order approximations of spectral graph convolutions and hence are unable to capture higher-order relational information between nodes. Probabilistic Graphical Models form another class of models that provide rich flexibility in incorporating such relational information but are limited by inefficient approximate inference algorithms at higher order. In this paper, we propose to combine these approaches to learn better node and graph representations. First, we derive an efficient approximate sum-product loopy belief propagation inference algorithm for higher-order PGMs. We then embed the message passing updates into a neural network to provide the inductive bias of the inference algorithm in end-to-end learning. This gives us a model that is flexible enough to accommodate domain knowledge while maintaining the computational advantage. We further propose methods for constructing higher-order factors that are conditioned on node and edge features and share parameters wherever necessary. Our experimental evaluation shows that our model indeed captures higher-order information, substantially outperforming state-of-the-art $k$-order graph neural networks in molecular datasets. 
	\end{abstract}
	
	\section{Introduction}
	Graph representation learning~\citep{hamilton2017representation,hamilton2017inductive} has gained more attention of researchers in recent years since much of the data available is structured and can be easily represented with graphs.
	The aim of representation learning on graphs is to encode structural information of graphs in order to learn better and richer node embeddings. These embeddings can then be used for downstream tasks like node classification, link prediction, graph classification and regression.
	
	In recent years, Graph Neural Networks (GNNs)~\citep{kipf2016semi,defferrard2016convolutional,gilmer2017neural,wu2020comprehensive} have shown remarkable success in learning better representations for graph structured data. GNNs learn node representations by iteratively passing messages between nodes within their neighbourhood and updating the node embeddings based on the messages received. Though successful, these models are limited by the first order approximations they make in aggregating information from the neighbouring nodes; graphs often exhibit complex substructures which cannot be captured by only pairwise modeling. Many graphs exhibit repeating substructures like motifs, and nodes satisfy higher-order constraints e.g., valence constraints in molecular data~\citep{wu2018moleculenet,agarwal2006higher}. 
	We can learn better node representations if we can design better message passing schemes that have sufficient representational power to capture such higher-order structures and constraints. 
	
	Node interactions have also been modeled with Probabilistic Graphical Models (PGMs)~\citep{wainwright2008graphical, koller2009probabilistic} wherein nodes are seen as random variables with a factorized joint distribution defined over them. Research focus in such models has been to find approximate inference algorithms to compute node marginals. There is rich literature of theoretically grounded PGM inference algorithms to find the state of a node given its statistical relations with other variable nodes in the graph. Knowledge of such inference algorithms can provide good inductive bias if it can be encoded in the neural network. Inductive bias lets the network favour certain solutions over others and if the bias is indeed consistent with the target task, it helps in better generalization~\cite{battaglia2018relational}. So, while we use deep learning methods to learn representations in an end-to-end setting, better generalization may be achievable for some tasks with the inductive bias of classical inference algorithms. Unfortunately, these approximate inference algorithms become inefficient at higher-order.
	
	One such well-known algorithm to find approximate node marginals is Loopy Belief Propagation (LBP)~\citep{murphy2013loopy,pearl2014probabilistic}. In this paper, we propose to leverage sum-product LBP to formulate the message passing updates and build a better graph representation learning model. To this end, we derive an efficient message passing algorithm based on LBP with arbitrary higher-order factors, with the assumption that the higher-order factors are of low rank and can be represented as a mixture of small number of rank-1 tensors. The derived message  passing updates only need two operations, matrix multiplication and Hadamard product and their complexity grows linearly with the number of variables in the factor.
	
	Further, we show how the message passing updates can be represented in a neural network module and unroll the inference as a computational graph. We allow the messages to be arbitrary real valued vectors (instead of being constrained to be positive as in LBP) and treat the messages as latent vectors in a network; the latent vectors produced by the network can then be used for the learning the target task through end-to-end training. We refer to the process of unrolling the algorithm, relaxing some of the constraints, and using the resulting network as a component for end-to-end learning as \emph{neuralizing} the algorithm. With LBP, this gives us a neural network message passing scheme, where higher-order factors can be added based on domain knowledge. 
	
	We focus on molecular data as a case-study. High-order information is likely useful in inference problems on molecular data but modeling the high-order interactions is challenging; nodes and edges are typed and the variable number of neighbours of each node makes fixed sized factors inappropriate. We show how to construct higher-order variable sized factors that share parameters through the use of edge and node features, allowing our model to be applied not only to molecular data, but also to other similar datasets where the graphs may have typed nodes and edges,  with different nodes having different degrees.  Experiments on two challenging large molecular datasets (QM9 and Alchemy) show that our model outperforms other recently proposed $k$-order GNNs substantially. 
	
	\section{Related work} 
	There have been significant number of works which have tried to incorporate inference algorithms in deep networks~\cite{zheng2015conditional,chen2015learning, lin2016efficient, lin2015deeply,tamar2016value,karkus2017qmdp,xu2017scene}. A large number of these works focus on learning pairwise potentials of graphical models on top of CNN to capture relational structures among their outputs for structured prediction. These methods are largely focused on modeling for a specific task like semantic segmentation~\citep{zheng2015conditional,lin2016efficient}, scene graph prediction~\citep{xu2017scene}, and image denoising~\citep{wu2016deep}. Further, ~\citep{tamar2016value,karkus2017qmdp} represent planning algorithms as neural network layers in sequential decision making tasks. 
	
	The inspiration for our low-rank formulation of factors comes from~\citep{wrigley2017tensor} where higher-order potentials were decomposed to derive an efficient sampling based junction-tree algorithm. Further, significant number of low-rank bilinear models have seen successful application in visual question-answering task~\citep{kim2016hadamard,kim2018bilinear,yu2017multi}. Similar low-rank methods were used in restricted Boltzmann machines as well for 3-way energy functions in~\cite{memisevic2007unsupervised,memisevic2010learning}.  Our formulation of higher-order factors as tensor decompositions can be seen as a generalization of low-rank bilinear pooling in~\citep{kim2016hadamard}.
	
	There has been some recent work on extending the graph convolutional neural networks to hyper-graphs in order to capture higher-order information~\citep{feng2019hypergraph,yadati2019hypergcn, jiang2019dynamic,zhang2019hyper}. ~\citet{feng2019hypergraph,jiang2019dynamic} used clique-expansion of hyper-edges to extend convolutional operation to hyper-graphs. Such modeling is equivalent to decomposing an hyper-edge to a set of pairwise edges. Similar approximation is applied in ~\citep{yadati2019hypergcn} where the number of pairwise edges added are reduced and are linearly dependent on the size of the hyperedge. Although, these methods operate on hyper-graphs, effectively the hyper-graphs are reduced to graphs with pairwise edges. 
	
	Recently, ~\citet{zhang2019factor} proposed a graph network expressive enough to represent max-product belief propagation. We work with the sum-product algorithm instead and further develop higher-order factors that can be flexibly applied to graphs containing nodes of different degrees. Further,~\citet{morris2019weisfeiler} and ~\citet{maron2019provably} used Weisfeiler Lehman (WL) graph isomorphism tests to construct increasingly powerful GNNs. They proposed models of message passing to capture higher order structures and compared their expressiveness with higher-order WL tests. As these models are theoretically shown to be more expressive in capturing higher-order information, we set these models as benchmarks and experimentally compare our model with the $k$-order GNNs.
	
	\section{Preliminaries}
	In this section, we briefly review two concepts central to our approach, low rank tensor decomposition and the sum product form of the loopy belief propagation algorithm.
	
	\subsection{Tensor decompositions}
	Tensors are generalizations of matrices to higher dimensions. A order-$m$ tensor $\mathbf{T}$ is an element in $\mathbb{R}^{N_1\times N_2\cdots\times N_m}$ with $N_k$ possible values in $k^{th}$ dimension for $k \in \{1,2,\dots ,m\}$. Tensor rank decompositions provide succinct representation of tensors. 
	In CANDECOMP / PARAFAC (CP) decomposition, a tensor $\mathbf{T}$ can be represented as a linear combination of outer products of vectors as
	\begin{equation}
	\mathbf{T} = \sum_{r=1}^R \lambda_r w_{r,1} \otimes w_{r,2} \otimes \cdots \otimes w_{r,m}
	\label{eq:tensor}
	\end{equation}
	where $\lambda_r\in\mathbb{R}$, $w_{r,k} \in \mathbb{R}^{N_k}$, $\otimes$ is the outer product operator i.e. $\mathbf{T}(i_1,i_2\dots,i_m)=\sum_{r=1}^R  \lambda_r w_{r,1}^{i_1} w_{r,2}^{i_2} \cdots w_{r,m}^{i_m}$, and the term $w_{r,1} \otimes w_{r,2} \otimes \cdots \otimes w_{r,m}$ is a rank-1 tensor. The scalar coefficients $\lambda_r$ can optionally be absorbed into  $\{w_{r,k}\}$. The smallest $R$ for which an exact $R$-term decomposition exists is the rank of tensor $\mathbf{T}$ and the decomposition (\ref{eq:tensor}) is its $R$-rank approximation. With this compact representation, an exponentially large tensor $\mathbf{T}$ with ${N_1\times N_2\cdots\times N_m}$ entries can be represented with $R$ vectors for each variable in $\mathbf{T}$, i.e. with a total of $R(N_1+N_2+\dots+N_m)$ parameters. More information about tensor decompositions can be found in ~\citep{kolda2009tensor,rabanser2017introduction}. 
	
	\subsection{Graphical models and Loopy belief propagation}
	Consider  a  hyper-graph $\mathcal{G} =  (\mathcal{V},\mathcal{E})$ with vertices $\mathcal{V}=\{1,\dots,n\}$ and hyperedges $\mathcal{E} \in 2^\mathcal{V}$. An undirected graphical model ($\mathcal{G}$,$\mathcal{F}$) associates every  vertex $i \in \mathcal{V}$ 
	with  a  discrete random  variable $x_i \in X$ and every hyperedge $a \in \mathcal{E}$ with a non-negative potential function $f_a \in \mathcal{F}$ such that the joint distribution of variables factorizes over $\mathcal{F}$ as
	\begin{align}
	P(x_1, x_2\dots,x_n) = \frac{1}{Z} \prod_{f_a \in \mathcal{F}} f_a(X_a)
	&& Z=\sum_{X}\prod_{f_a \in \mathcal{F}} f_a(X_a)
	\label{eq:pgm}
	\end{align}
	where $X_a=[x_i:i \in a]$ and $Z$ is the normalizing constant. Without loss of generality, we assume all variables can take $d$ values and consequently $f_a(X_a) \in \mathbb{R}^{d^{|X_a|}}$. A factor graph is a convenient data structure used to represent such distributions. It is a bipartite graph whose edges connect variable nodes to factor nodes such that every $x_i$ is connected to $f_a$ if $x_i \in X_a$. Throughout the paper, we index variables with $\{i,j,\dots\}$ and factors with $\{a,b,\dots\}$.
	
	Loopy belief propagation (LBP)~\citep{pearl2014probabilistic,murphy2013loopy} is the procedure of computing approximate marginals $p(x_i)$ at each node $x_i$ on a factor graph by sending messages between factor and variables nodes. Essentially, LBP starts by initializing two kinds of messages, factor-to-node $m_{a\rightarrow i}(x_i)$ and node-to-factor $m_{i\rightarrow a}(x_i)$. Messages are a function of the variable in the receiving node updated with the following recursive equations,
	\begin{equation}
	m_{i\rightarrow a }(x_i) = \prod_{c \in  N(i) \setminus \{a\}} m_{c\rightarrow i} (x_i)
	\label{eq:msg_i_a} 
	\end{equation}
	
	\begin{equation}
	m_{a\rightarrow i}(x_i) = \sum_{X_a \setminus \{x_i\}} f_a(X_a) \prod_{j \in  N(a) \setminus \{i\}} m_{j\rightarrow a} (x_j)
	\label{eq:msg_a_i}
	\end{equation}
	where $N(i)$ is the set of neighbours of $i$ and $m_{a\rightarrow i}(x_i) \in \mathbb{R}^d$. After sufficient number of iterations, the belief of variables is computed by
	\begin{equation}
	b_i(x_i) = f_i(x_i) \prod_{a \in N(i)} m_{a\rightarrow i} (x_i).
	\label{eq:bel_i}
	\end{equation}
	
	\section{Proposed Method}
	\subsection{Low Rank Loopy Belief Propagation}
	Consider a factor $f_a(X_a)$ over $|X_a|=n_a$ variables with $X_a = [x_1, x_2 \dots, x_{n_a}]$. Since, $f_a(X_a)$ is a tensor in $\mathbb{R}^{d^{n_a}}$, it can be represented as a sum of $R$ fully factored terms in CP decomposition form (\ref{eq:tensor}). 
	\begin{equation}
	f_a(X_a) = \sum_{r=1}^R \lambda_r w_{a_{r,1}} \otimes w_{a_{r,2}} \otimes \cdots \otimes w_{a_{r,n_a}}
	\label{eq:fa_tensor}
	\end{equation}
	where $w_{a_{r,j}} \in \mathbb{R}^d$. This representation is efficient if $f_a$ is of low-rank i.e., $R \ll d^{n_a}$. We posit that such a low-rank representation is often a good approximation of $f_a$ in practice; the method is likely to be useful when this assumption holds. 
	
	Absorbing $\lambda_r$ in~(\ref{eq:fa_tensor}) into weights $\{ w_{a_{r,j}} \}$ and substituting in (\ref{eq:msg_a_i}), we have
	\begin{equation}
	m_{a\rightarrow i}(x_i) = \sum_{X_a \setminus \{x_i\}} \Big(\sum_{r=1}^R w_{a_{r,1}} \otimes w_{a_{r,2}} \otimes \cdots \otimes w_{a_{r,{n_a}}} \Big) \prod_{j \in  N(a) \setminus \{i\}} m_{j\rightarrow a} (x_j)
	\label{eq:lbp_1}
	\end{equation}
	
	\begin{equation}
	m_{a\rightarrow i}(x_i) = \sum_{X_a \setminus \{x_i\}} \sum_{r=1}^R \big( w_{a_{r,1}} m_{1\rightarrow a} (x_{1})\big) \otimes \cdots w_{a_{r,i}} \cdots \otimes  \big(w_{a_{r,n_a}} m_{{n_a}\rightarrow a}(x_{n_a})\big) 
	\label{eq:lbp_2}
	\end{equation}
	Now we can marginalize out a variable in the same manner as if $f_a$ was fully factorized. We simply push the outer summation inside, distribute and separately evaluate it over each of the univariate products $\Big(w_{a_{r,j}} m_{j\rightarrow a} (x_{j})\Big)$. This gives us
	\begin{equation}
	m_{a\rightarrow i}(x_i) = \sum_{r=1}^R w_{a_{r,i}}  \Big(\sum_{x_1} w_{a_{r,1}} m_{1\rightarrow a} (x_{1})\Big) \cdots \Big( \sum_{x_{n_a}} w_{a_{r,{n_a}}} m_{{n_a}\rightarrow a} (x_{n_a})\Big) 
	\label{eq:lbp_3}
	\end{equation}
	
	\begin{equation}
	m_{a\rightarrow i}(x_i) = \sum_{r=1}^R  w_{a_{r,i}}  \gamma_{a_{r,1}} \cdots \gamma_{a_{r,{n_a}}}
	\label{eq:lbp_4}
	\end{equation}
	where $\sum_{x_j} w_{a_{r,j}}  m_{j\rightarrow a}(x_j) = w_{a_{r}}^T m_{j\rightarrow a} = \gamma_{a_{r,j}} \in \mathbb{R}$.  
	If we stack $R$ weight vectors for each variable as $W_{a_i} = [ w_{a_{1,i}},w_{a_{2,i}},\dots w_{a_{R,i}}] \in \mathbb{R}^{d\times R}$ and $\boldsymbol{\gamma}_{a_j} =[\gamma_{a_{1,j}},\gamma_{a_{2,j}}, \dots \gamma_{a_{R,j}}]^T\in \mathbb{R}^{R\times 1}$, we can rewrite the (\ref{eq:lbp_4}) in matrix form as 
	\begin{equation}
	m_{a\rightarrow i} = W_{a_i} \big[\boldsymbol{\gamma}_{a_1} \odot \boldsymbol{\gamma}_{a_2} \dots \odot \boldsymbol{\gamma}_{a_{n_a}} \Big]
	\label{eq:lbp_5}
	\end{equation}
	where $\boldsymbol{\gamma}_{a_j} = W_{a_j}^{T} m_{j\rightarrow a}$ and $\odot$ is the Hadamard product or element-wise multiplication.
	
	
	Now we have new message passing updates for the loopy belief propagation algorithm,
	\begin{equation}
	m_{a\rightarrow i} = W_{a_i} \Big[\big(W_{a_1}^{T} m_{1\rightarrow a} \big) \odot \big(W_{a_2}^{T} m_{2\rightarrow a} \big)  \dots \odot \big(W_{a_{n_a}}^{T} m_{{n_a}\rightarrow a} \big) \Big]
	\label{eq:lbp_msg_a}
	\end{equation}
	
	\begin{equation}
	m_{i\rightarrow a } = \bigodot_{c \in  N(i) \setminus \{a\}} m_{c\rightarrow i} 
	\label{eq:lbp_msg_b} 
	\end{equation}
	Belief update is simple. For the variable $x_i$, we project the messages from the other variables sharing a factor with $x_i$ to $\mathbb{R}^R$, perform elementwise multiplication and then project the product back to $\mathbb{R}^d$. Thereafter, multiply such messages from all the factors $x_i$ is connected to get the updated belief of $x_i$. 
	
	Clearly, the computational complexity of the message updates grows linearly with the addition of variables to factors and thereby the algorithm is efficent enough to run with higher-order factors.
	
	\subsection{Neuralizing Low Rank LBP}
	To learn better node and consequently, graph representations in an end-to-end setting, we seek to neuralize the Low Rank LBP algorithm by writing the message passing updates as a functionally equivalent  neural  network module and \textit{unroll} the inference algorithm as a computational graph. We further replace the positive message vectors in LBP with unconstrained real valued hidden latent vector representations initialized from a feature extractor network.  Depending on the problem, the feature extractor can be modeled to either capture zeroth or first order information. Thereafter, the message updates can follow the higher-order factor graph structure. In our experiments on molecular datasets, we use MPNN \citep{gilmer2017neural} as the feature extractor network and on an optical character recognition dataset, we use CNN as a base feature extractor. 
	
	
	\begin{wrapfigure}{r}{.4\textwidth}
		\begin{minipage}{\linewidth}
			\centering
			\includegraphics[scale=0.13]{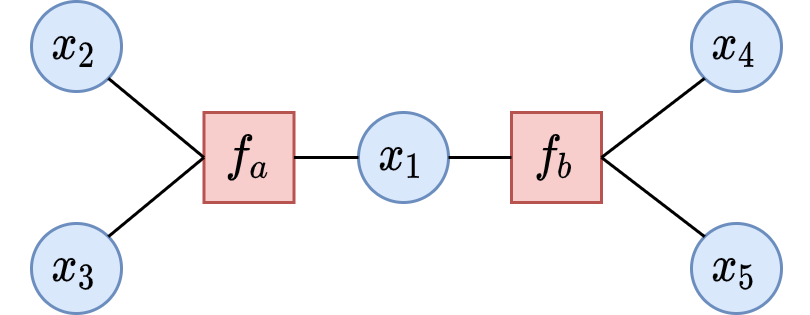}
			\caption*{(a)}
			\label{fig:5a}\par\vfill
			\includegraphics[scale=0.09]{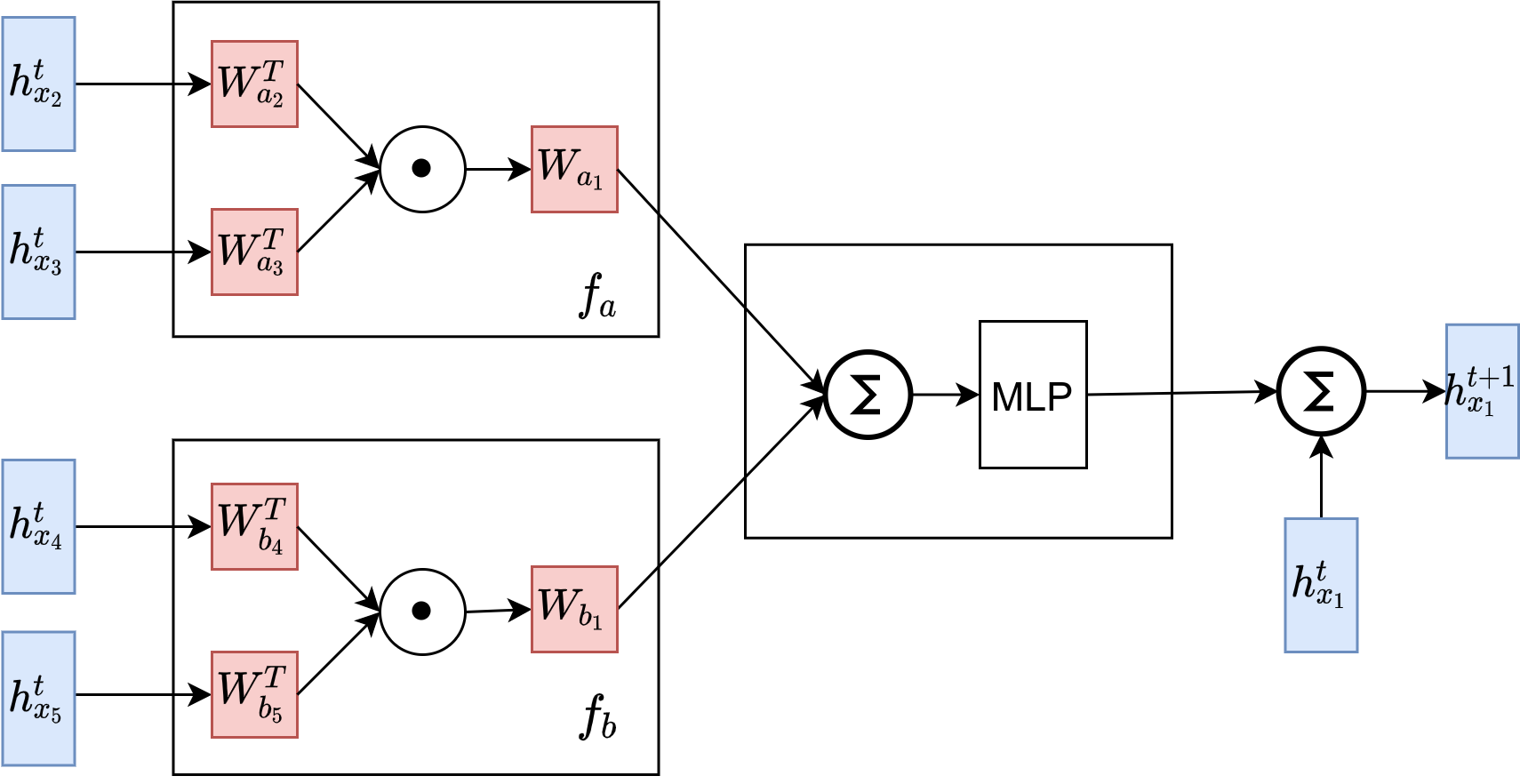}
			\caption*{(b)}
			\label{fig:5b}
		\end{minipage}
		\caption{(a) Factor graph with $x_1$ and its neighbours. (b) Message passing update for node $x_1$}\label{fig:5}	
	\end{wrapfigure}
	The LBP algorithm contains multiplication of several terms leading to numerical instability due to overflow and underflow errors. It has been shown that normalization does not alter the final beliefs in LBP ~\citep{pearl2014probabilistic} and hence standard practice of handling these numerical errors is normalization of messages after every iteration of the algorithm. In our experiments, we tried methods like normalization after multiplication in equation~(\ref{eq:lbp_msg_b}) and addition of log-normalized messages from higher-order factors.  Unfortunately, we found that such methods quickly lead to sub-optimal local minima, 
	and the network fails to learn discriminative representations. Empirically, we found that replacing the multiplication operation in the message update function of (\ref{eq:lbp_msg_b}) with summation followed by an MLP layer with relu activation works well in enabling stable learning of useful representations.
	We note that the network no longer represents (\ref{eq:lbp_msg_b}) exactly, although it retains the remaining structure of the algorithm, capturing most of its inductive bias. We leave it to future work to handle the numerical issues related to exactly representing (\ref{eq:lbp_msg_b}) during training.
	
	We now describe the Low Rank LBP network. Consider a Graph $\mathcal{G}=(\mathcal{V},\mathcal{E})$ with node and edge features. Let $\mathcal{F}$ be a factor graph defined on $\mathcal{G}$. 
	Let $h_{x_i}^t$ be the hidden state at node $x_i$ at iteration $t$. Then, we define the Low Rank LBP network with the following message passing update:
	\begin{equation}
	h_{x_i}^{t+1} = h_{x_i}^{t} + MLP \Bigg(\sum_{a \in  N(i)} 
	W_{a_i} \Big[\bigodot_{j \in  N(a) \setminus \{i\}}W_{a_j}^{T} h_{x_j}^t \Big] \Bigg)
	\label{eq:lbpnet_msg} 
	\end{equation}
	where $N(i)$ are the set of factors adjacent to $x_i$, $N(a)$ the set of variables adjacent to factor $a$ and $\{W_{a_j}\}$ are learnable parameters. 
	We add the previous hidden state with the new message as residual connection to improve learning. After $T$ iterations, we can use a readout function on the learnt node representations for prediction tasks on nodes or the whole graph. 
	
	A factor is described by the set of matrices $\{W_{a_j}\}$ in (\ref{eq:lbpnet_msg}) i.e. it has as many $W_{a_j}$'s as the number of variables adjacent to it in the factor graph. We further relax this constraint and maintain $2|\{W_{a_j}\}|$ matrices at each factor, one set is used for transformation before Hadamard product and the other set after the product. 
	This increases the representative power of the network while still being able to represent  (\ref{eq:lbpnet_msg}) as the extra set of matrices can be learnt to be same as the first set.
	
	\section{Models}
	\subsection{Modeling handwritten character sequences with higher-order factors}
	A sequences is one of the simplest graph structures. In a handwritten character sequence, the previous few characters are likely to contain useful information for predicting the next character. We consider one of the simplest higher order model: a higher order factor is placed at each node in the sequence. We use this model to experimentally study the scaling behaviour of our algorithm.
	
	\subsection{Modeling molecular data with higher-order factors}\label{modeling_molecular_data}
	Molecular data has several properties needed for an effective study of higher-order representation learning models of graphs. A molecule is a graph with atoms as nodes and bonds that exist between atoms as edges. Higher-order information is present in the form of valence constraints of atoms which determine number of bonds they can form. Molecules often contain subgraphs, also called functional groups (e.g., OH, CHO etc.), which are identifiable and significant in determining their functionality and geometry. Furthermore, molecules can have highly variable connectivity, arbitrary size and variety of 3D molecular shapes. 
	They come with broad range of chemical properties to predict with large heterogeneity in their structures~\citep{wu2018moleculenet}. 
	Any graph learning model should be powerful enough to capture such higher-order structures and dependencies in order to learn highly discriminative representations. 
	Thereby, we focus on molecular data to study the effectiveness of the proposed model and show its modeling flexibility in incorporating domain knowledge in constructing the factors. 
	
	In (\ref{eq:lbpnet_msg}), the set of weights $\{W_{a_j}\}$ are separate for all factors and within a factor, separate for all variables connected to the factor. 
	This gives much freedom in modeling to leverage domain knowledge in order to share some of these parameters and be able to work with large graphs as well. Given a molecular graph, we discuss possible ways of constructing higher-order factors, conditioning and sharing of parameters. 
	
	One way to capture higher-order information is to add a factor for every node in the molecule connecting that node (we will call this node \textit{central atom} in that factor) and its neighbours to the factor. Then factor weights can be shared by conditioning on the  following 
	\begin{itemize}[{leftmargin=*}]
		\item \textbf{Central atom type (CAT):} Weights within the factor are shared but different factors share parameters only if they have the same central atom type. In this case, we maintain only as many weight matrices as the number of atom types.
		\item \textbf{Bond type (BT):} Weights are shared if the bond type between the central atom and its neighbour is same. In this case, we maintain as many weight matrices as the number of bond types. 
		\item \textbf{Central atom and bond type (CABT):} Weights are shared if both the central atom type and bond type are same. Number of weight matrices needed is the product of number of atom and bond types.
		\item \textbf{Central atom, bond and neighbour atom type (CABTA):} Weights are shared if the bond type and the atom types of atoms sharing the bond are same. 
	\end{itemize}
	Do note that most molecular datasets have small number of atom types and bond types. Further, if we have continuous edge features or other features types such as strings, then each weight matrix $W_{a_j}$ can be learnt by end-to-end training using an MLP with node and edge features as inputs and the weight matrix as output. This shows the proposed model of message passing is flexible and provides sufficient freedom in modeling.
	
	\section{Experiments}
	We first evaluate Low Rank Belief Propagation (LRBP) network on recognizing handwriting characters in a sequence, a task where higher order dependencies is likely to be useful. We use this task to better understand the scaling behaviour of the model.
	Further, to validate the effectiveness of LRBP-net in capturing higher-order information from more general graph structured data, we report results on two molecular datasets and compare with other higher-order GNNs that capture such information as well -- the methods we are comparing with also give the state-of-the-art performances. 
	
	\subsection{Optical character recognition data}
	\begin{figure}
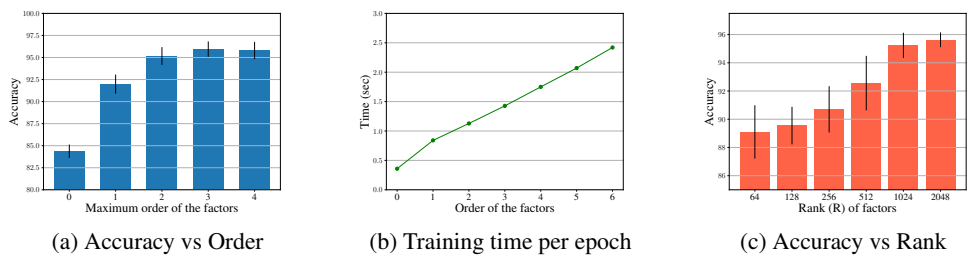

		\centering
		\begin{minipage}{0.32\textwidth}
			\centering
			\scalebox{0.2}{\input{ocr_a.pgf}}
			\subcaption{Accuracy vs Order}
			\label{fig:graph_a}
		\end{minipage}
		\begin{minipage}{0.32\textwidth}
			\centering
			\scalebox{0.2}{\input{ocr_b.pgf}}
			\subcaption{Training time per epoch}
			\label{fig:graph_b}
		\end{minipage}
		\begin{minipage}{0.32\textwidth}
			\centering
			\scalebox{0.2}{\input{ocr_c.pgf}}
			\subcaption{Accuracy vs Rank}
			\label{fig:graph_c}
		\end{minipage}
		\caption{Handwriting character Recognition}\label{fig:graph}	
	\end{figure}
	To study the properties of LRBP network, we used the handwriting recognition data set~\citep{taskar2004max}, originally collected by ~\citep{kassel1995comparison}. 
	The dataset consists of a subset of $\sim 6100$ handwritten words with the average length of $\sim 8$ characters. The words are segmented into characters and each character is rasterized into an image of 16 by 8 binary pixels. The dataset is available in 10 folds with each fold containing $\sim 6000$ characters. The task is to recognise characters by leveraging the neighbourhood of characters within the context of word. Since, the words come from a small vocabulary, there is a strong higher-order correlation present in the dataset. We follow the 10-fold cross-validation setup in~\citep{taskar2004max} with 1 fold for training and 9 folds for testing and report averaged results.
	In our framework, each individual character image is a node $x_i$ with the label $y_i \in \{a,b,\dots z\}$. Depending on the order, $x_i$ can share a factor with other character nodes $x_j$ within the same word. We evaluate the performance of LRBP-net by varying order and rank of the factors.
	
	\textbf{Implementation: }We use 3 standard convolution and a fully connected layer as the base the feature extractor to get the zero-th order feature of 512 dimension. We then run 3 iterations of higher-order message passing followed by a classifier. We fix the rank of all factors at 1024 and share  parameters between factors with same order. We train for 50 epochs with a learning rate of 0.001.
	
	Results in Figure~(\ref{fig:graph_a}) show strong improvements as maximum order of factors used is successively increased before saturating at 4th-order and above. To evaluate the efficiency of (\ref{eq:lbpnet_msg}), we analysed the computation time as we vary the order of factors used. Figure~(\ref{fig:graph_b}) shows that the training time per epoch grows almost linearly with the order of factors.
	
	To evaluate the effect of rank of the factors, we fixed the zero-th order feature dimension to 64 and used factors up to order-3. We then ran three message passing iterations by varying the rank of the factors from 64 to 2048. Results in Figure~(\ref{fig:graph_c}) show consistent improvement in performance with the increasing rank. We suggest the rank should be at least twice the message hidden vector dimension.
	
	\subsection{Molecular data}
	\textbf{Datasets:} We evaluate our model on QM9~\citep{ruddigkeit2012enumeration,ramakrishnan2014quantum} and Alchemy~\citep{chen2019alchemy} datasets on the task of regression on 12 quantum-mechanical properties. QM9 is composed of 134K drug-like organic molecules with sizes varying from 4 to 29 atoms per molecule. Atoms belong to one of the 5 types, Hydrogen (H), Carbon (C), Oxygen (O), Nitrogen(N), and Fluorine (F) and each molecule may contain up to 9 heavy (non-Hydrogen) atoms. Nodes come with discrete features along with their 3D positions. We follow the standard 80:10:10 random split for training, validation and testing. 
	
	Alchemy~\citep{chen2019alchemy} is a recently released more challenging dataset of 119K organic molecules comprising of 7 atom-types H, C, N, O, F, S (Sulphur) and Cl (Chlorine) with up to 14 heavy atoms. These molecules are screened for being more likely to be useful for medicinal chemistry based on functional group and complexity. Furthermore, we follow the split based on molecule size where almost all training set contains molecules with 9 or 10 heavy atoms while the validation and test set contain molecules with 11 or 12 heavy atoms. As quantum-mechanical properties depend on the molecule size, this split tests how well the model can generalize to heavier molecules. The regression targets are same as in QM9 dataset.
	
	\textbf{Implementation:} We plug our message passing scheme within Message Passing Neural Network (MPNN) framework~\citep{gilmer2017neural} with the  standard implementation provided by Pytorch-geometric~\citep{fey2019fast}. MPNN has Edge-conv~\citep{gilmer2017neural} as message function, GRU~\citep{chung2014empirical} as update function followed by a set2set~\citep{vinyals2015order} model as readout function for the graph. We run 3 iterations of MPNN message passing scheme followed by 3 iterations of higher-order LRBP message passing described in (\ref{eq:lbpnet_msg}). We then combine the MPNN output from the third iteration and the LRBP output with concatenation followed by the set2set readout function. For LRBP module, we set the hidden vector ($h_{x_i}$) dimension to 64 and the rank $R$ of factors to 512. We use Adam optimizer initialized with learning rate of $1e^{-3}$. All other hyperparameters are maintained as is provided by Pytorch-geometric implementation. All targets are normalized and are trained with the absolute error loss for 200 epochs with batch size of 64. Refer supplementary material for more details.
	\begin{table}[tbp]
		\scriptsize
		\caption{Graph Regression results on QM9 dataset}
		\label{table-qm9}
		\centering
		\begin{tabular}{lrrrrrrrrr}
			\toprule
			& & \multicolumn{4}{c}{Joint training of targets} & \multicolumn{4}{c}{Separate training of targets} \\
			\cmidrule(lr){3-6} \cmidrule(lr){7-10}
			Target & MPNN & 123-GNN & PPGNN & LRBP & Gain(\%) & 123-GNN & PPGNN & LRBP & Gain(\%)  \\
			\midrule
			$\mu$ & 0.358                & 0.407            & 0.231                & \textbf{0.092}              & 60.19                & 0.476                & 0.0934               & \textbf{0.0688}               & 26.33                  \\
			$\alpha$& 0.89                 & 0.334              & 0.382                & \textbf{0.183}     & 45.20                & 0.27                 & 0.318                & \textbf{0.14028}              & 48.04                \\
			$\epsilon_{homo}$& 0.00541              & 0.07807              & 0.00276              & \textbf{0.00199}     & 27.89                & 0.0037               & 0.00174              & \textbf{0.00168}             & 2.98                  \\
			$\epsilon_{lumo}$& 0.00623              & 0.08492              & 0.00287              & \textbf{0.00205}     & 28.57                & 0.00351              & 0.0021               & \textbf{0.0016}          & 22.95                 \\
			$\Delta_\epsilon$& 0.0066               & 0.10939              & 0.00406              & \textbf{0.00279}     & 31.28                & 0.0048               & 0.0029               & \textbf{0.0024}             & 15.51                \\
			$\langle R^2\rangle$& 28.5                 & 22.83            & 16.07                & \textbf{2.81}     & 82.53                & 22.9                 & 3.78                 & \textbf{1.41}              & 62.69                \\
			$ZPVE$& 0.00216              & 0.01129              & 0.00064              & \textbf{0.00018}     & 71.87                & 0.00019              & 0.000399             & \textbf{0.00010}             & 47.37                 \\
			$U_0$& 2.05                 & 8.910              & 0.234                & \textbf{0.0907}     & 61.25                & 0.0427               & 0.022                & \textbf{0.0163}            & 25.90                \\
			$U$& 2                    & 8.910              & 0.234                & \textbf{0.0907}     & 61.23                & 0.111                & 0.0504               & \textbf{0.0139}               & 72.42                \\
			$H$& 2.02                 & 8.910              & 0.229                & \textbf{0.0906}     & 60.42                & 0.0419               & 0.0294               & \textbf{0.0175}              & 40.47                \\
			$G$& 2.02                 & 8.910              & 0.238                & \textbf{0.0907}     & 61.91                & 0.0469               & 0.024                & \textbf{0.0134}               & 44.16               \\
			$C_v$& 0.42                 & 0.1184              & 0.184                & \textbf{0.084}     & 29.05                & 0.0944               & 0.144                & \textbf{0.0552}              & 41.52                 \\
		\end{tabular}
	\end{table}
	
	\textbf{Results:}
	\begin{wraptable}{r}{.7\textwidth}
		\scriptsize
		\centering
		\caption{Graph Regression results on Alchemy dataset}
		\label{table-alchemy}
		\begin{tabular}{lrrrrrrr}
			\toprule
			\multirow{2}{*}{Target} & \multirow{2}{*}{MPNN*} & \multirow{2}{*}{LRBP} & \multirow{2}{*}{Gain(\%)} & \multicolumn{4}{c}{LRBP Network Ablation Models} \\
			\cmidrule(lr){5-8} 
			&  &  & & CAT & BT & CABT & CABTA \\
			\midrule
			$\mu$& \textbf{0.1026}  & 0.1041 & -1.41   & 0.1091  & 0.1041  & 0.1092  & 0.1233 \\
			$\alpha$& 0.0557 & \textbf{0.0451}  & 19.05  & 0.0473  & 0.0451 & \textbf{0.0446}  & 0.0449 \\
			$\epsilon_{homo}$& 0.1151  & \textbf{0.1004} & 12.74   & 0.1065   & 0.1004  & \textbf{0.1001}  & 0.1007  \\
			$\epsilon_{lumo}$& 0.0817  & \textbf{0.0664} & 18.74 & 0.0712 & \textbf{0.0664}  & 0.0685  & 0.0696  \\
			$\Delta_\epsilon$& 0.0832   & \textbf{0.0691} & 16.88 & 0.0739   & \textbf{0.0691} & 0.0703   & 0.0720    \\
			$\langle R^2 \rangle$& 0.0271   & \textbf{0.0099} 	& 63.47   & 0.0099     & 0.0099  & \textbf{0.0094}   & 0.0120 \\
			$ZPVE$& 0.0259   & \textbf{0.0115} 	& 55.42 & 0.0116   & 0.0115      & \textbf{0.0108}  & 0.0140   \\
			$U_0$& 0.0131    & \textbf{0.0044}	& 65.80   & \textbf{0.0042}  & 0.0044    & 0.0046    & 0.0054    \\
			$U$& 0.0131    & \textbf{0.0044} 	& 65.90   & \textbf{0.0041}  & 0.0044    & 0.0046   & 0.0054   \\
			$H$& 0.0130     & \textbf{0.0044}	& 65.77  & \textbf{0.0042}     & 0.0044    & 0.0046    & 0.0054 \\
			$G$& 0.0130     & \textbf{0.0044}	& 65.77  & \textbf{0.0042}     & 0.0044  & 0.0046   & 0.0054  \\
			$C_v$& 0.0559  & \textbf{0.0481}	&13.93    & \textbf{0.0472}  & 0.0481   & 0.0488   & 0.0502  \\
			\midrule
			MAE & 0.0499  &	\textbf{0.0394}		& 21.18   & 0.04115  & \textbf{0.0394}  & 0.0400  & 0.0424     \\
		\end{tabular}
	\end{wraptable}
	For QM9 dataset, following~\cite{maron2019provably} we report Mean Absolute Error (MAE) in two settings, one where all targets are jointly trained and the other where each target is separately trained. Factors are constructed as described in Section~\ref{modeling_molecular_data} with factor weights conditioned on central atom and bond type (CABT). 123-GNN~\citep{morris2019weisfeiler} and PPGNN~\citep{maron2019provably} are the $k$-order methods which capture higher-order information with PPGNN having the state-of-the-art score in QM9 dataset. Table~\ref{table-qm9} shows that LRBP-net outperforms PPGNN by significant margin in all the targets under both the settings. Furthermore, the margin of improvement indicates that much of the higher-order information was not sufficiently captured by the $k$-order GNNs.
	
	For Alchemy dataset, following~\cite{chen2019alchemy} we report MAE on jointly trained normalized targets and compare with MPNN which was the best performing model in the benchmark results~\cite{chen2019alchemy}. We use the validation set to select among the models in Section~\ref{modeling_molecular_data}. As LRBP-net is built on MPNN as the backend network, the margin of improvement in Table~\ref{table-alchemy} is mainly because of higher-order message passing module. We also did an ablation study using models described in Section~\ref{modeling_molecular_data}. Results indicate that conditioning of parameters on either central atom or edge type helps most. Conditioning in these ways help capture most of the higher-order information centered around an atom (node). For the CABTA method, there is slight decrease in performance which is likely caused by the large parameter size in the model.
	Collectively, the ablation results suggest that major improvements are coming from the message passing scheme itself since conditioning on only bond types (BT) seems to be sufficient for better performance.
	
	\section{Conclusion}
	In this paper, we present an efficient Low-rank sum-product belief propagation procedure for inference in factor graphs where the complexity of message updates grows linearly with the number of variables in the factor. Further, we neuralize the message passing updates to learn better node representations with end-to-end training. This gives us a fairly simple, flexible, powerful and efficient message passing scheme for representation learning of graph data where higher-order information is present. We validate the proposed model with experiments on molecular data where it outperforms other state-of-the-art $k$-order GNNs substantially. 
	
	\newpage
	\section*{Broader Impact}
	We propose a new efficient representation learning technique for graph structured data. It is a general algorithm to learn better graph representations and would impact all the areas where data can be represented in the form of sparse graphs. Particularly, we show better performance on molecular data and hence our method will likely be helpful in applications which require predictions on molecular data. We foresee our method to be helpful in applications like drug discovery where new drugs can be discovered. 
	
	These graph based learning techniques can also be used to learn behaviour of communities. Such technologies may be misused by persons with mal-intent. Such outcomes can be curbed by increased research into privacy technologies and appropriate regulations by governmental organizations. 

	\bibliographystyle{apalike}
	\bibliography{sample.bib}

	\newpage
	\begin{center}
		\LARGE \textbf{Appendix}
	\end{center}
	
	\section{Improvement over MPNN when distance information is excluded}
	In the main paper, we reported the improved performance of LRBP-net over MPNN on Alchemy dataset. The LRBP-net is built on MPNN as backend feature extractor. Therefore, its improved performance is the result of capturing dependencies not modeled by MPNN. This begs the question: can higher-order message passing capture more information when the backend MPNN module is further constrained. For this, we limit the input to MPNN module and see if LRBP-net can further improve its gain with respect to MPNN. 
	
	One of the main reasons for the superior performance of MPNN on molecular datasets is that it can capture the 3D geometric structure of the molecule~\cite{chen2019alchemy,gilmer2017neural}. MPNN is provided with edge features which include bond type and spatial distance between the pair of atoms. Further, it operates on a complete graph where extra \emph{virtual edges} are added between every pair of atoms with no bond. The edge feature for such \emph{virtual edges} contains the spatial distance between the pair of atoms. Consequently, MPNN can capture 3D geometric structure of the molecule with such a complete graph. 
	
	In the following experiments, we evaluate whether higher-order message passing can help capture structure of the molecule in the absence of pairwise distance edge features. 
	We do include 3D atom positions in the node features and hence the information about the geometric structure of the molecule is indirectly provided. Based on the pairwise distance feature, we divide the experimental setup in three categories .
	\begin{itemize}[{leftmargin=*}]
		\item \textbf{Sparse graph without distance:} Input graph is a sparse graph i.e., edge exists only if bond exists between atoms, with edge features containing the bond type without the  distance between the pair of atoms. 
		\item \textbf{Sparse graph with distance:} Input graph is a sparse graph with edge features containing bond type and distance between the pair of atoms.
		\item \textbf{Complete graph with distance:} Input graph is a complete graph with extra \emph{virtual edges} containing distance information between pair of atoms in the edge. This setup is the standard MPNN model.
	\end{itemize}
	In all three cases, regardless of input to the MPNN module, the higher order message passing works only on the sparse graph.
	\begin{table}[ht]
		\footnotesize
		\caption{Comparison  of LRBP-net with MPNN on Alchemy dataset with/without distance information}
		\label{table-suppl-alchemy}
		\centering
		\begin{tabular}{lccccccccc}
			\toprule
			\multirow{2}{*}{Target} & 
			\multicolumn{3}{c}{Sparse graph without distance}& 
			\multicolumn{3}{c}{Sparse graph with distance}& 
			\multicolumn{3}{c}{Complete graph with distance} \\
			\cmidrule(lr){2-4} \cmidrule(lr){5-7} \cmidrule(lr){8-10}
			& MPNN & LRBP & Gain(\%) & MPNN & LRBP & Gain(\%) & MPNN & LRBP & Gain(\%) \\
			\toprule
			$\mu$                 & 0.3546              & 0.3071              & 13.39           & 0.3655              & 0.3012              & 17.60           & 0.1026     & 0.1041               & -1.41         \\
			$\alpha$              & 0.1971              & 0.0873              & 55.68  & 0.1639              & 0.0888              & 45.80         & 0.0558              & 0.0451              & 19.06          \\
			$\epsilon_{homo}$     & 0.1723              & 0.1281              & 25.66   & 0.1539              & 0.1220              & 20.73         & 0.1151              & 0.1005              & 12.75           \\
			$\epsilon_{lumo}$     & 0.1280              & 0.0905              & 29.31     & 0.1120              & 0.0847              & 24.37         & 0.0817              & 0.0664     & 18.74          \\
			$\Delta_{\epsilon}$   & 0.1266              & 0.0902              & 28.75     & 0.1084              & 0.0854              & 21.18         & 0.0832              & 0.0692     & 16.88          \\
			$\langle R^2 \rangle$ & 0.1439              & 0.0629     & 56.31     & 0.2926              & 0.0629     & 78.50        & 0.0271              & 0.0099     & 63.47          \\
			$ZPVE$                & 0.1388              & 0.0412              & 70.28    & 0.1020              & 0.0399              & 60.80         & 0.0259              & 0.0115              & 55.42          \\
			$U_0$                 & 0.0806              & 0.0206              & 74.39         & 0.0606              & 0.0236              & 61.05         & 0.0131               & 0.0045              & 65.80          \\
			$U$                   & 0.0806              & 0.0206              & 74.37      & 0.0606              & 0.0236              & 61.05         & 0.0131              & 0.0045              & 65.90          \\
			$H$                   & 0.0806              & 0.0209              & 74.07                & 0.0606              & 0.0236              & 60.98          & 0.0131              & 0.0045              & 65.77          \\
			$G$                   & 0.0806              & 0.0208              & 74.13           & 0.0606              & 0.0236              & 61.02         & 0.0131              & 0.0045              & 65.77          \\
			$C_v$                 & 0.2177              & 0.0913              & 58.04                  & 0.1729              & 0.0931              & 46.13         & 0.0559               & 0.0481              & 13.93          \\
			\toprule
			MAE & 0.1501         & 0.0818             & \textbf{45.50}   & 0.1428         & 0.0810             & \textbf{43.24}         & 0.0499        & 0.0394        & \textbf{21.18}         
		\end{tabular}
	\end{table}
	
	Results in Table~\ref{table-suppl-alchemy} shows that the margin of improvement of LRBP-net over MPNN is significantly higher when pairwise distance feature is not included in the graph. This suggests MPNN is not able to sufficiently capture the 3D molecular shape in both the cases where sparse graph is used, In such scenarios, capturing higher-order structures with LRBP module is more helpful in reducing the MAE. Furthermore, results of both the cases are similar when sparse graph is used and inclusion of pairwise distance in the edge feature does not lead to significant performance gains. Following this, it can be inferred that bond types are indicative of distance between the atoms as well. 
	
	\section{Results of ablation models on QM9 dataset:}
	In the main paper, we considered ablation models of LRBP-net based on conditioning of factor parameters. These models are CAT (central atom type), BT (bond type), CABT (central atom and bond type) and CABTA (central atom, bond type and neighbouring atom type). We reported results on Alchemy dataset where we found that conditioning on bond type was sufficient for good performance. To further verify the results, we evaluate the ablation models on QM9 dataset.
	
	Results in Table~\ref{table-qm9_sup_abl} show that unlike Alchemy dataset, CABT model performs better in almost all the targets on QM9 dataset. This perhaps suggests that in QM9 dataset, higher-order constraints are more centered around the atom and are better captured by having separate parameters for different central atom types. Collectively, the ablation study on QM9 and Alchemy datasets suggests that conditioning the parameters on neighbouring atom type (CABTA) is not helpful and only increases the paramter size. It is sufficient if edge type and central atom type information is directly captured in the model. 
	\begin{table}[thbp]
		\footnotesize
		\caption{Ablation models regression results on QM9 dataset}
		\label{table-qm9_sup_abl}
		\centering
		\begin{tabular}{lrrrr}
			\toprule
			\multirow{2}{*}{Target} &  
			\multicolumn{4}{c}{LRBP Network Ablation Models} \\
			\cmidrule(lr){2-5} 
			& CAT & BT & CABT & CABTA \\
			\midrule
			$\mu$ & \textbf{0.088}     &0.095               & 0.092              & 0.097                 \\
			$\alpha$ &  0.192               & 0.204              & \textbf{0.183}     & 0.185                \\
			$\epsilon_{homo}$        & 0.0021               &0.00217              &  \textbf{0.00199}     & 0.00203               \\
			$\epsilon_{lumo}$ &  0.00212              &0.0022               &  \textbf{0.00205}     & 0.00208               \\
			$\Delta_{\epsilon}$  & 0.0029               &0.00301              &  \textbf{0.00279}     & 0.00286               \\
			$\langle R^2 \rangle$     & 2.88              & 3.01              &\textbf{2.81}     & 3.17                \\
			$ZPVE$ &  0.00021              & 0.00021              &  \textbf{0.00018}     & 0.00022                \\
			$U_0$ & 0.1283              & 0.0995              &   \textbf{0.0907}     & 0.1649            \\
			$U$ &  0.1283              & 0.1029              &  \textbf{0.0907}     & 0.1650               \\
			$H$ & 0.1283              & 0.1008              & \textbf{0.0906}     & 0.1649                \\
			$G$ &  0.1283              & 0.1011              &  \textbf{0.0906}     & 0.1651                \\
			$C_v$ &  0.0868              & 0.0900              & \textbf{0.0847}     & 0.0872                 \\
			\midrule
			MAE & 0.3141 & 0.3183 &  \textbf{0.2947} & 0.3512 
		\end{tabular}
	\end{table}
	
	\section{Computational cost:}
	To demonstrate the computational efficiency of LRBP-net, we report the training time per epoch for both the large molecular datasets and compare with the MPNN model. We run all our experiments on NVidia RTX-2080Ti GPU with Pytorch framework~\cite{paszke2017automatic} and its pytorch-geometric extension~\cite{fey2019fast}.
	
	Table~\ref{table-suppl-time} shows that the computational cost of LRBP-net is not more than twice that of MPNN on average. Note that LRBP-net is built upon MPNN with 3 iterations of MPNN message passing followed by 3 iterations of higher-order LRBP message passing. Therefore, the amortized computational cost of an iteration of higher-order message passing is similar to the cost of an iteration of MPNN message passing.
	\begin{table}[h]
		\footnotesize
		\caption{Training time per epoch (sec)}
		\label{table-suppl-time}
		\centering
		\begin{tabular}{lcc}
			\toprule
			& MPNN & LRBP-net  \\
			\#Iterations & 3 & 6\\
			\midrule
			QM9 & 148.61 & 312.81 \\
			Alchemy & 149.60 & 271.64 \\
			\toprule
			Average & 149.10 & 292.22
		\end{tabular}
	\end{table}
	
	\begin{table}[t!]
		\footnotesize
		\caption{Target properties of QM9 and Alchemy datasets}
		\label{table-qm9-supl}
		\centering
		\begin{tabular}{lll}
			\toprule
			Property & Unit & Description  \\
			\midrule
			$\mu$& Debye              & Dipole moment                 \\
			$\alpha$& Bohr$^3$ & Isotropic polarizability            \\
			$\epsilon_{homo}$& Hartree & Energy of highest occupied molecular orbital (HOMO)       \\
			$\epsilon_{lumo}$& Hartree & Energy of lowest occupied molecular orbital (LUMO)    \\
			$\Delta_{\epsilon}$ & Hartree & Gap, difference between LUMO and LUMO  \\
			$\langle R^2 \rangle$& Bohr$^2$ & Electronic spatial extent \\
			$ZPVE$& Hartree & Zero point vibrational energy            \\
			$U_0$& Hartree & Internal energy at 0 K      \\
			$U$& Hartree & Internal energy at 298.15 K        \\
			$H$& Hartree & Enthalpy at 298.15 K         \\
			$G$& Hartree & Free energy at 298.15 K      \\
			$C_v$& cal/(mol K)  & Heat capacity at 298.15 K     \\
			\toprule 
		\end{tabular}
	\end{table}

\end{document}